\documentclass[letterpaper]{article} % DO NOT CHANGE THIS
\usepackage[submission]{aaai25}  % DO NOT CHANGE THIS
\usepackage{times}  % DO NOT CHANGE THIS
\usepackage{helvet}  % DO NOT CHANGE THIS
\usepackage{courier}  % DO NOT CHANGE THIS
\usepackage[hyphens]{url}  % DO NOT CHANGE THIS
\usepackage{graphicx} % DO NOT CHANGE THIS
\usepackage{natbib}  % DO NOT CHANGE THIS AND DO NOT ADD ANY OPTIONS TO IT
\usepackage{caption} % DO NOT CHANGE THIS AND DO NOT ADD ANY OPTIONS TO IT
\usepackage{algorithm}
\usepackage{algorithmic}

\usepackage{booktabs}%
\usepackage{amsmath,amssymb,amsfonts}%
\usepackage{todonotes}
\usepackage{pifont}
\usepackage{times}
\usepackage{graphicx}
\usepackage{subcaption}
\usepackage{colortbl}
\usepackage{array}
\usepackage[T1]{fontenc}
\usepackage[utf8]{inputenc}
\usepackage{microtype}
\usepackage{inconsolata}
\usepackage{enumitem}
\usepackage{caption}
\usepackage{algorithm}
\usepackage{algorithmic}

\usepackage{booktabs}%
\usepackage{amsmath,amssymb,amsfonts,amsthm}%
\usepackage{todonotes}
\usepackage{pifont}

\usepackage{newfloat}
\usepackage{listings}
\DeclareCaptionStyle{ruled}{labelfont=normalfont,labelsep=colon,strut=off} % DO NOT CHANGE THIS
\floatstyle{ruled}
\newfloat{listing}{tb}{lst}{}
\floatname{listing}{Listing}
\title{Fairness at Every Intersection: Uncovering and Mitigating Intersectional Biases in Multimodal Clinical Predictions}
\author{
    Kishore Sampath\textsuperscript{\rm 1},Ayaazuddin Mohammad\textsuperscript{\rm 1},
    Resmi Ramachandranpillai\textsuperscript{\rm 2},
}
\affiliations{
    \textsuperscript{\rm 1}Northeastern University, USA\\
    \textsuperscript{\rm 2}Institute for Experiential AI, Northeastern University, USA\\
    \{sampath.ki, mohammad.ay, r.ramachandranpillai\}@northeastern.edu
}
\usepackage{bibentry}
\begin{document}

\maketitle

\begin{abstract}
Biases in automated clinical decision-making using Electronic Healthcare Records (EHR) impose significant disparities in patient care and treatment outcomes. Conventional approaches have primarily focused on bias mitigation strategies stemming from single attributes, overlooking intersectional subgroups - \textit{groups formed across various demographic intersections (such as race, gender, ethnicity, etc.)}. Rendering single-attribute mitigation strategies to intersectional subgroups becomes statistically irrelevant due to the varying distribution and bias patterns across these subgroups. The multimodal nature of EHR - \textit{data from various sources such as combinations of text, time series, tabular, events, and images} -  adds another layer of complexity as the influence on minority groups may fluctuate across modalities. In this paper, we take the initial steps to uncover potential intersectional biases in predictions by sourcing extensive multimodal datasets, MIMIC-Eye\footnote{The only multimodal dataset available for research that encloses patients' information from all modalities (including x-rays).} and MIMIC-IV ED, and propose mitigation at the intersectional subgroup level. We perform and benchmark downstream tasks and bias evaluation on the datasets by learning a unified text representation from multimodal sources, harnessing the enormous capabilities of the pre-trained clinical Language Models (LM), MedBERT, Clinical BERT, and Clinical BioBERT. 
%We then conduct an in-depth intersectional bias analysis and propose bias mitigation tailored to the intersectional subgroup level. 
Our findings indicate that the proposed sub-group-specific bias mitigation is robust across different datasets, subgroups, and embeddings, demonstrating effectiveness in addressing intersectional biases in multimodal settings.
    
\end{abstract}

% Uncomment the following to link to your code, datasets, an extended version, or similar.

\section{Introduction}
\begin{figure}[hbt!]
    \centering
    \includegraphics[width=0.48\textwidth, height=5.5cm]{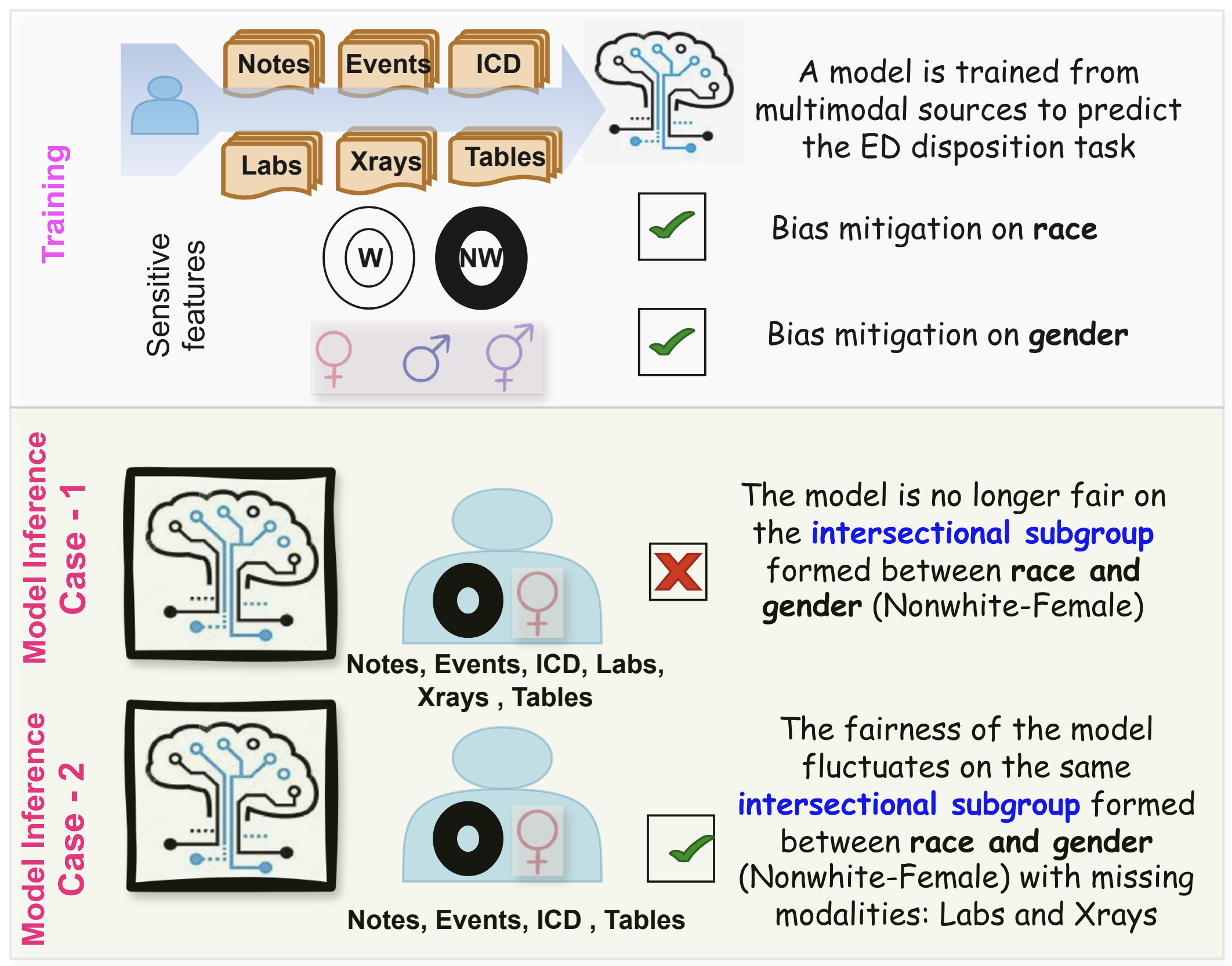} % Replace with your image file name
    \caption{Intersectional biases in multimodal settings; Bias mitigation focused at single attribute is no longer valid for demographic intersections (case 1). The nature of multimodal data adds another layer of complexity in terms of fairness fluctuations (case 2). }
    \label{fig:interm}
\end{figure}

Electronic Healthcare Records (EHR) have revolutionized the optimization and delivery of healthcare, providing a digital database of patient information that enhances the efficiency and effectiveness of clinical decision-making \cite{seymour2012electronic}. Automating clinical decision-making using EHR is not without significant challenges, particularly with the perpetuation and amplification of biases. These biases can adversely impact the quality and equity of care provided to diverse patient populations and their intersections \cite{pivovarov2014identifying, boyd2023equity, rouzrokh2022mitigating, banerjee2023shortcuts}.

Existing measures for intersectional biases in healthcare \cite{ogungbe2019systematic, okoro2022intersectional} overlooked the multimodal dimensionality \cite{turan2019challenges, amal2022use, kline2022multimodal, pang2021multi}
%To efficiently enhance the prediction performance of AI-powered clinical decision-making, 
as learning from diverse sources affects data coherence, feature interoperability, and introduces varying bias patterns. Some efforts have been made to learn unified feature representation for disease diagnosis generation \cite{niu2024ehr} and emergency department predictions \cite{lee2024multimodal}. Figure \ref{fig:interm} shows various challenges in dealing with intersectional biases in multimodal learning.  The challenge of intersectionality becomes even more pronounced in multimodal settings as the fairness value fluctuates with different modality information. Moreover, single-attribute based fairness mitigation strategies become irrelevant as the subgroup densities vary significantly when considering demographic intersections, and balancing those sub-group populations in multimodal training (such as through counterfactual generation and data reweighting) is not practically feasible \cite{alabdulmohsin2024clip}. In Figure \ref{fig:interm}, training is conducted with bias mitigation applied separately for race and gender. During model inference (Case 1), the model exhibits a significant bias toward the intersectional patient group formed by race and gender, utilizing information from all available data sources. However, in Case 2, the same model does not show bias toward the intersectional patient group when X-rays and lab results are excluded. It is thus important to note that modality-specific bias mitigation will fail in this scenario as different modalities exhibit different bias patterns and intensities \cite{yan2020mitigating}\footnote{The authors have proposed data balancing and adversarial strategies, but the multimodality dimension is limited.} and the biases in multimodal model may not correlate with those found in individual modalities.  

Motivated by the aforementioned challenges, we propose \textbf{Subgroup-specific Discrimination Aware Ensembling (SDAE), an ensemble of multimodal classifier models tailored to address layered complexities of demographic intersectional biases}. \textbf{To capture vast knowledge of patient's health conditions from multimodal sources, we rely on the extensive power of pre-trained LMs: MedBERT \cite{rasmy2021med}, Clinical BERT, and Bio-Clinical BERT \cite{alsentzer2019publicly} for generating clinically relevant embeddings in the form of texts}. \textbf{Employing pre-trained clinical LMs to generate embeddings has an added advantage of facilitating learning with small-scale data, like MIMIC-Eye, harnessing the potential of rich and high dimensional clinical knowledge acquired through pre-training}. To accommodate these clinical LMs, we learn the \textbf{representations in unified textual format from all diverse multimodal sources}. We perform evaluations using MIMIC-Eye \cite{hsieh2023mimic} and MIMIC-IV ED \cite{johnson2020mimic} and our method is robust to any tasks/domains involving demographic intersections and multimodal learning. 

Note that the MIMIC-IV ED does not exhibit any severe disparities in terms of the intersections. Therefore, to test the robustness of our method, we created a biased sample\footnote{More details and original bias evaluation are given in the supplementary file, Section 5} from MIMIC-IV ED and used it for all the downstream tasks and bias analysis. 

%The significance of our work (Figure \ref{fig:interm}) can address the following issues: (

%The EHR systems based on Machine Learning (ML) or Deep Learning (DL) models center around information from unimodal sources \cite{mullenbach2018explainable, wang2018joint, qiao2019mnn, niu2021label, harutyunyan2019multitask}, leaving the vast knowledge from multimodal sources — encompassing clinical text, events or time series, radiology images, and other data types — underexplored. 

 %An instance from MIMIC-eye describes this scenario in Figure \ref{fig:interm}. When excluding time series data(right),  the performance drops by 10\% despite showing fair outcomes across all the demographic intersections in terms of Worst case Parity (WP)\footnote{WP is a metric used for measuring the intersectional biases \cite{ghosh2021characterizing}}. The fairness improvement can be justified by the influence of time series data which contributes to Asian females but decreasing the overall performance.
 %We advocate for choosing modalities that contribute to the overall prediction performance after careful examination on individual modality contributions on the predictions and subsequently design bias mitigation strategies considering intersectional subgroups rather than binary sensitive settings. .

\setlength{\parskip}{0pt}
\textbf{Contributions}. Our major contributions address the complex challenges of intersectional biases in multimodal healthcare. Additionally, the MIMIC-Eye dataset has not yet been explored for downstream tasks and biases due to its inherent complex nature of multimodality which amplifies the overall impact of our work. 
%Our work takes the initial steps in exploring these tasks using multimodal information and performing a comprehensive bias analysis focusing on intersectionality and proposing mitigation. 
The major contributions can be summarized as:
\setlength{\parskip}{0pt}
\begin{enumerate}[nosep]%[itemsep=-6.5pt]
\item We propose bias mitigation through multimodal subgroup-specific ensemble classifiers. We design a post-process bias mitigation customized at the sensitive subgroup intersections (Section \ref{sec:SDAE}).
    \item We benchmark downstream tasks on MIMIC-Eye analyzing modality-wise contributions to prediction accuracy and comparing these results to the overall prediction performance achieved through multimodal learning (Section \ref{sec:tasks}). %We conduct a detailed study on the unimodal contributions to the predictions and compare it with the multimodality performance (Section \ref{sec:bias}).
    
    \item In-depth intersectional bias analysis on the benchmarking tasks on MIMIC-Eye (Section \ref{sec:bias}).  
    \item Empirical analysis of single attribute post-process bias mitigation approaches and their effect on intersectionality. We show that existing bias mitigation approaches cannot be extended to settings combining intersectionality and multimodality (Section \ref{sec:bias}). 
\end{enumerate}

\section{Background}
\label{sec:background}
We consider binary and multitask (binary) classifications. The data, $D=\{X_n\}_{i=1}^n$
of each instance contains information from clinical notes ($EHR\_ {notes}_n$), Xrays/radiology images ($EHR\_ {Xray}_n$), clinical events ($EHR\_ {events}_n$), laboratory test results ($EHR\_ {lab}_n$), and from the tabular structure ($EHR\_ {structured}_n$). The sensitive features $S$ are from the $EHR\_ {structured}_n$. The data instance $X$ encloses information from all the modalities and can be written as:

%\vspace{-18pt}
\begin{equation}
\label{a}
X_i =
\begin{pmatrix}
\begin{array}{l}
\text{$EHR\_{notes}_i$}, \text{$EHR\_{events}_i$}, \\
\text{$EHR\_{lab}_i$}, \text{$EHR\_{Xray}_i$}, \\\text{$EHR\_{structured}_i$}
\end{array}
\end{pmatrix}
\end{equation}

The downstream prediction tasks on each $X_i$ contain a ground truth label $Y \in \{0,1\}$. Since we consider healthcare settings, we don't differentiate between favorable and unfavorable classes, as it depends on the underlying task at hand. We assume multiple sensitive settings, $s=\{s_1, s_2,....s_m\}$, where $m$ represents the total number of sensitive attributes in the dataset. Then we define subgroups,  $\text{SG}_{gs_1\times gs_2\times...gs_m}={gs_1} \cap {gs_2} \cap... \cap {gs_m}$ who belong to group $gs_1$ through $gs_m$ with marginal sensitive attributes $s_1, s_2,....s_m$ .  For example, if $s=\{gender, race\}$, $gender \in \{male, female\}$, and $ \text race \in \{white,non-white\}$, then we will have 4 subgroups, $\text{SG}$, (i) $\{female, white\}$, (ii) $\{female, non{-}white\}$, (iii) $\{male, white\}$, and (iv) $\{male, non{-}white\}$, $|{\text{SG}}|=4$. 

\subsection{Algorithmic Fairness }
\textbf{Definition 1: Demographic Parity (DP)} \cite{barocas2016big} - Let $f$ be a function $f: X \rightarrow \hat Y, \hat Y=\{0,1\}$ for binary classification, and let $S$, be a sensitive attribute, the function $f$ satisfies DP if:

\vspace{-15pt}
\begin{multline}
P[f(x)=1 \mid x \in S] = P\left[f(x)=1 \right], \forall s \in S
\end{multline}

where $x$ denotes an instance of $X$ and $P[.]$ denotes the probability of an instance. 

\vspace{5pt}
\textbf{Definition 2:
Equal Opportunity or True Positive Rate Parity (TPR)} for a binary prediction $\hat{Y}$ and a member subgroup $A$, is satisfied if:
% \begin{equation}
% \begin{aligned}
% P(\hat{Y} = 1| A \in \text{SG}_i, Y = 1)=\\ & P(\hat{Y} = 1| A \in \text{SG}_j, Y = 1) \\
% & \forall i, j \in |\text{SG}|, \, i \neq j
% \end{aligned}
% \end{equation}
%\vspace{-15pt}
\begin{multline}
P(\hat{Y} = 1 \mid A \in \text{SG}_i, Y = 1) = \\
P(\hat{Y} = 1 \mid A \in \text{SG}_j, Y = 1) \\
\forall i, j \in |\text{SG}|, \, i \neq j
\end{multline}

\subsection{Intersectional Fairness}
For defining intersectional fairness in this work, we follow \cite{ghosh2021characterizing}. 

The worst-case parity (WP) of measuring intersectional DP can be framed as follows \cite{ghosh2021characterizing}:
\vspace{-5pt}
\begin{equation}
\label{Eq3}
  \text{WP(DP)} = \frac{\min\{P(\hat{Y} \mid A \in \text{SG}i), \forall i \in |\text{SG}|\}}{\max\{P(\hat{Y} \mid A \in \text{SG}j), \forall j \in |\text{SG}|\}},
\end{equation}

where $i\neq j$. Similarly, the worst-case parity of intersectional TPR can be defined as:
% \vspace{-5pt}
\begin{equation}
\label{Eq4}
\text{WP(TPR)} = 
\frac{
    \min\{P(\hat{Y} = 1 \mid A \in \text{SG}_i, Y = 1)}
{
    \max\{P(\hat{Y} = 1 \mid A \in \text{SG}_j, Y = 1)},
\end{equation}
where $\forall i \in |\text{SG}|,  \forall j \in |\text{SG}|$, and $i\neq j$. For a definition of algorithmic fairness, $\mathcal U(S, \hat{Y})$ (DP or TPR) on the predictions $\hat{Y}$, the WP can be calculated by taking the min-max ratio of values from the given set of subgroups. A value closer to 1 guarantees fair outcomes and we follow the $80\%$ rule. 
% Using \ref{Eq3} the Disparate Impact of intersectional subgroups \cite{ghosh2021characterizing} can be measured by:
 %\begin{equation}
 %\label{Eq2}
% \begin{aligned}
%DI = \min \left( \frac{P(\hat{Y}  \mid A \in SGi)}{P(\hat{Y} | A \in SGj)}\right) \\
%& \forall i,j \in |SG|, i\neq j 
 %\end{aligned}
%\end{equation}
%In our experiments, we consider equation \ref{Eq1} and \ref{Eq2} for the definition of subgroup fairness $\mathcal U(SG, \hat{Y})$.

\textit{\textbf{Remark}: We recommend choosing a fairness measure based on the context of the application in healthcare. The DP measure focuses solely on positive outcomes and does not consider ground truth values. However, in situations where outcomes are conditioned on the patient's acute index - for example, if all acute index values are similar - we suggest DP conditioned on acute index levels. Otherwise, we recommend using TPR as it helps in detecting underdiagnosed biases.}

\section{Methodology}
\begin{figure*}[t]
    \centering
    \includegraphics[width=0.88\textwidth,height=7cm]{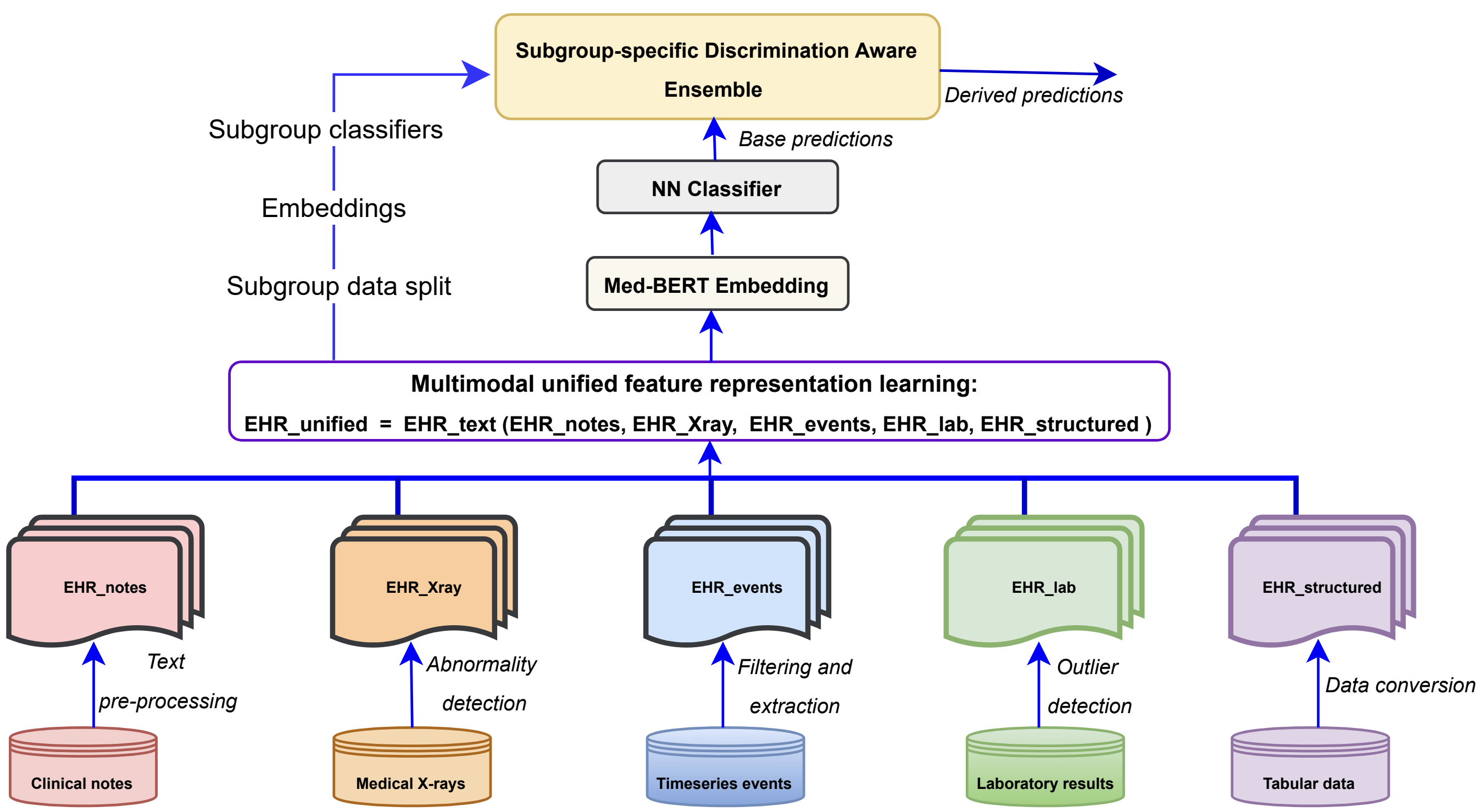} 
    \caption{The proposed architecture}
    \label{fig:example}
\end{figure*}

In this section, we describe the proposed workflow (Supplementary file, Section 1) and method in detail (Figure \ref{fig:example}). 
\subsection{Learning Unified Text Representations from Multimodal Data}

The first step in our proposed work is to gather information from multimodal sources. To capture the information into a unified feature representation, we extend the approach described in \cite{lee2024multimodal}, where textual representations of tabular EHR features have been learned. However, it overlooked the heterogeneous multimodal nature of the EHR data, which we address by extending the concept of textual representation learning to the remaining modalities. 

We perform pre-processing to remove noises from the clinical texts to obtain $EHR\_ {notes}$. For generating radiology reports from the chest x-rays, we follow \cite{tanida2023interactive}. Given a chest X-ray, we run it through an object detection model to extract 29 regions of interest and run a binary image classifier to find any abnormalities. A Language model has been employed for creating the abnormality descriptions followed by a post-processing step to create $EHR\_ {Xray}$. For clinical events, we adopt a filtering mechanism to remove repeated events which are subsequently described in textual form, $EHR\_ {events}$. For laboratory results (with time stamps) we follow the box-plot outlier detection \cite{rousseeuw2018anomaly} to locate abnormal portions which are then textualized as $EHR\_ {lab}$. Finally, the tabular representations are textualized $(EHR\_ {structured})$ using the method described in \cite{lee2024multimodal}.

The unified multimodal feature representation learning  can be written as:
\vspace{-5pt}
\begin{multline}
    EHR\_ {\text{unified}}(X_i) = EHR\_{\text{text}} (X_i) \\
    \forall X_i, i \in \{1, 2, \ldots, n\}
\end{multline}
Here, $EHR\_ {\text{unified}}(X_i)$ denotes the unified textual embedding for a patient $X_i$ from multimodal sources and $EHR\_{\text{text}}$ denotes the resultant textual representation.

\subsection{Embedding and Prediction}
One could argue that with the emergence of Large Language Models like GPT, training a model from scratch using embeddings is no longer necessary. However, this argument has been refuted in \cite{lee2024multimodal}.
To generate embeddings for the unified EHR concepts in the $EHR\_ {\text{unified}}$ textual representation and enable the downstream models to learn interoperable and cross-modal features, we employ clinical LMs: Med-BERT, ClinicalBERT, and Bio-clinicalBERT, which centers on the BERT framework containing contextualized embeddings of the EHR dataset of $28,490,650$ patients. Note that, the unified representation now contains information from all the sources unlike \cite{lee2024multimodal}. During the training stage, tokenized data from $EHR\_{\text{unified}}$ is fed into the clinical LMs. These encoders produce embeddings that capture information from multimodal sources of a patient's EHR in rich and high-dimensional vector representations. We keep the encoders frozen at this stage and update the weights of the subsequent prediction models dedicated to tasks. The mathematical modeling can be represented as :
\begin{equation}
\label{Eq 6}
\begin{aligned}
    \vec{E} = \text{Clinical LMs}(c_i),
    \end{aligned}\\
\end{equation}
where, $\forall c_i \in Tokenizer(EHR\_{\text{unified}})$.
Once we generate the embeddings, the final output is sent to the subsequent classifiers for predictions. We train a self-attention neural network classifier to align the cross-modal features followed by fully connected layers as this combination will enhance the predictive capabilities of the model \cite{lee2024multimodal}. This can be represented as :
\begin{equation}
    \vec{E}_{\text{attention}} = \text{SelfAttention}(\vec{E});    
\end{equation}
\begin{equation}
    \vec{E}_{fc} = \text{ReLU}(FC(\vec{E}_{\text{attention}}))
\end{equation}
The classifier output can then be modeled:
\begin{equation}
\hat{Y} =\min_{ {\text{loss}\_{task}}} (\text{Classifier}(\vec{E}_{fc})),
\end{equation}
where ${\text{loss}\_{task}}$ is the loss tailored for the individual binary predictions. After obtaining the task predictions, an extensive bias analysis has been done to uncover potential biases.

\subsection{Subgroup-specific Discrimination Aware Ensembling (SDAE) }
\label{sec:SDAE}
The final step is to design bias mitigation strategies considering intersectionality and multimodality. In this work, we consider post-process bias mitigation for two main reasons:
\begin{enumerate}[nosep]
    \item Data preprocessing approaches \cite{kamiran2012data} such as Reweighting and disparate impact remover \cite{feldman2015certifying} will not be feasible in multimodality as data sources are dispersed, making the identification of data-label pairs impractical. %This is true in our settings despite the unified text representation learning as the application of those strategies to text remains unattended.
    \item In-processing methods such as adversarial debiasing \cite{zhang2018mitigating} often fall short of completely optimizing an adversary \cite{zhu2021learning} and their effect on intersectional subgroups is an ongoing research area, which we leave for future study.
\end{enumerate}
Based on the above observations we consider methods that enforce fairness through post-processing. Our goal is to learn  $\mathcal U(\text{SG}, \hat{Y})$- enforced classifier models for the downstream binary and multitask prediction tasks trained on the multimodal patient records, where $\text{SG}$ is the set of intersectional subgroups as detailed in Section \ref{sec:background}. 

\textbf{Definition 3: Bias mitigation procedure} -  Let $\mathcal M$ be the base classifier, $\hat Y$ be the predictions of $\mathcal M$, then a bias mitigation procedure $\mathcal P$ is defined as a function that can be applied on the true predictions $\hat Y$ of $\mathcal M$ so that the resultant derived predictions $\hat Z$ of $\mathcal M$ is $\mathcal U(A, \hat{Z})-fair$, for all subgroup members,$ A \in \text{SG}$.

\setlength{\parskip}{0pt}
\textit{Training:} The proposed bias mitigation procedure $\mathcal{P}$ involves a post-processing approach through a series of steps to achieve derived predictions $\hat{Z}$. For this, we rely on bootstrap ensembling \cite{breiman1996bagging} and extend it to intersectional and multimodal settings. For mitigation, we build the procedure $\mathcal{P}$ on the unified feature representation $EHR\_{\text{unified}}$ and build multimodal models $\mathcal{M}^{M}$ corresponding to each split, $EHR\_{\text{unified}}^{M}$, where $M$ is the number of total models/dataset splits for subgroup pairs in $\text{SG}$. For example, if $\text{SG}$=$\{ $(i).(Female, White), (ii).(Female, Black), (iii).(Male, White), (iv).(Male, Black)$\}$, then the split will be based on the pairs in $\text{SG}$, such as $\{EHR\_{\text{unified}}^M \}$ = $\{ {EHR\_{\text{unified}}}^{i-ii}$, ${EHR\_{\text{unified}}}^{i-iii}$, ${EHR\_{\text{unified}}}^{i-iv}$, ${EHR\_{\text{unified}}}^{ii-iii}$, ${EHR\_{\text{unified}}}^{ii-iv}$, ${EHR\_{\text{unified}}}^{iii-iv} \}$, sum up to 6 splits enclosing multimodal subgroup-specific information. These splits are then embedded individually using Med-BERT and subgroup-specific classifiers are trained on the embeddings for predictions. The main idea behind this technique is that each classifier will learn the subgroup-specific features and contribute to the final ensembling stage. The models $\mathcal M^{i-j}$, where $i,j$ denotes that the model is trained on the split ${EHR\_{\text{unified}}}^{i-j}$. 

\textit{Prediction:} In prediction, each instance in the test/validation split is processed by models that correspond to the subgroup membership of that instance and the base multimodal classifier. We then form a Discrimination-Aware Ensemble \cite{kamiran2012decision} by exploiting the disagreement region of all the subgroup-specific classifiers. Relying solely on voting to calculate the score may result in many instances having identical score values making it challenging to decide which instances should be selected for class changes based on their scores, especially when the number of subgroups is small. To resolve this issue, we perform the following: (i) if all voters are in consensus, we assign the corresponding class label; otherwise, we compute a score $\eta$ for each instance $X_i$  as a linear combination of the classifier's probabilities $P^{\mathcal M^{i-j}}$ and votes $V^{i-j}$: 
\begin{equation}
  \begin{aligned}
    \text{If agreement,}\hspace{0.5cm} Z_i &= \bar{V}^{i-j}; \\
\text{Else,}\hspace{0.5cm} \eta_i &= h \cdot \bar{V}^{i-j} + (1-h) \cdot \bar{P}^{\mathcal{M}^{i-j}}; \\
Z_i &= 
\begin{cases} 
    1 & \text{if } \eta_i > \tau_{SG}, \\ 
    0 & \text{otherwise}
\end{cases}
    \end{aligned}
\label{eq.vote}
\end{equation}
where $\bar{V}^{i-j}$ is the winner (if not in consensus, it will be the ratio of the majority), $X_i$ carries a subgroup membership with models $\mathcal {M}^{i-j}$, $\bar{P}^{\mathcal M^{i-j}}$ is the average of the classifier probabilities, $Z_i$ is the derived prediction, and $\tau_{SG}$ (0.5 in our case) is a hyperparameter, that can be set to balance the fairness-accuracy trade-off for each of the subgroups. The parameter $h$ is introduced to prevent the predicted probabilities from overshadowing the $\bar{V}^{i-j}$ selection and can be set according to the number of models involved in the voting process as, $\frac{|M^{i-j}|-1}{|M^{i-j}|}$.
Note that, aggregating predictions, when all the models are in consensus will not have any significance on fairness. We exploit the disagreement region of individual predictors and use our ensembling strategy to improve fairness.

\section{Experiments}
\subsection{Downstream Tasks}
\label{sec:tasks}

\textbf{The MIMIC-Eye dataset} \cite{hsieh2023mimic}\footnote{https://physionet.org/content/mimic-eye-multimodal-datasets/1.0.0/} integrates various datasets associated with the Medical Information Mart for Intensive Care (MIMIC), offering a comprehensive collection of patient information. It includes medical images such as chest X-rays (available in both MIMIC CXR and MIMIC JPG formats), clinical data from the MIMIC IV Emergency Department (ED), detailed patient hospital journey records from MIMIC IV, and eye-tracking data capturing gaze information and pupil dilations. Detailed statistics of the dataset is given in the supplementary file, Section 2. We consider both MIMIC-Eye and MIMIC-IV ED for our experiments.

%To ensure data quality and consistency, preprocessing and data imputation were conducted and tailored to the analysis objectives. The dataset is organized based on patient information, with stay-IDs derived from patient records serving as the primary linkage between different modalities. In our research, each data point represents a specific patient stay, encompassing all relevant information for that stay. 

 %The MIMIC-Eye dataset was created to apply multimodal deep-learning models, but, no studies have yet utilized it for such purposes. This is the first to use this dataset in a multimodal deep learning context, with a comprehensive analysis of performance and biases, sets a precedent for future studies aiming to leverage the full potential of multimodal EHR in improving healthcare outcomes. 
 \begin{table}[]
    \centering
     \small
    \begin{tabular}{lccc}
   
        \toprule
        \textbf{Modality} & \textbf{Med} & \textbf{Clinical} & \textbf{Bio-} \\
        &\textbf{BERT}&\textbf{BERT}&\textbf{ClinicalBERT}\\
        \midrule
        Arrival+
        Triage+   &  &  &  \\
        Medrecon+  & 0.920  & 0.913  & 0.913 \\
        \hline
        Vitals    & 0.734 & 0.744 & 0.744 \\
         \hline
        Codes+  
        Pyxis     & 0.740 & 0.738 & 0.736\\
        \hline
        Xray      & 0.757 & 0.923 & 0.919 \\
        \midrule
        \textbf{Multimodal(+events)} & \textbf{0.925} & \textbf{0.741} & \textbf{0.913} \\
        \bottomrule
    \end{tabular}
    \caption{Multimodality analysis using F1 score and various BERT embeddings in MIMIC-Eye ED Disposition.}
    \label{tab:eddisp_performance}
\end{table}

\textbf{Benchmarking tasks}: In this study, we explore the downstream tasks to analyze the predictive performance. Specifically, we perform the binary classification and multitask classification \cite{lee2024multimodal}.
\begin{itemize}[nosep]
    \item ED Disposition - This involves predicting the post-visit destination of patients following their ED visit. This is a binary classification problem where label $1$ denotes patients who were admitted to ED and $0$ denotes patients who were discharged home. 
    \item ED Decompensation - It is a multitask binary classification and is used to predict three ED tasks simultaneously. The first task predicts the patient's discharge location (1:home, 0:not home). The second task predicts the need for an ICU. The third predicts patient mortality, specifically whether the patient will die during their hospital stay. 
\end{itemize}
 We use F1 \cite{hicks2022evaluation}
 %AUROC \cite{FAWCETT2006861}, and AUPRC \cite{saito2015precision} 
 to compare the performances of these tasks. The experimental settings are given in the supplementary file, Section 3. 

 \textbf{Multimodality selection}: We analyze the model's performance change when supplemented with modality across embeddings. The underlying idea is that a well-designed model trained from multiple sources should not exhibit performance degradation \cite{chen2024multimodal} \footnote{The ME-BEC dataset is not publicly available} when some information is missing at inference. 
%Some researchers nullify this and found that unimodal models perform better than their multimodal counterparts. 
%We link this to the modality underutilization studied in \cite{makino2023detecting}. When there are more modalities involved in the prediction, the models undergo shortcut learning where the models rely only on a subset of modalities. 
Analysis of individual performances of modalities against multimodal models can better assist in developing more robust models \cite{chen2024multimodal}. 
Following this observation, we train the downstream prediction models
on an increasing subset of modalities using multiple clinical LMs. We initiate the process with a multimodal model trained only on categorical
and numeric attributes, then incrementally add modalities from diagnoses, medications, orders, lab results, radiology reports, etc. Table \ref{tab:eddisp_performance} summarizes our analysis and the performance of multimodal models in the ED disposition tasks using MedBERT, clinicalBERT, and Bio-clinicalBERT embeddings. We provide the results of multitask learning in the supplementary file (Section 6). Based on the analysis, one can choose MedBERT over other embeddings as it better captures the multimodal interoperability contributing to an F1 score of 0.925 compared to other clinical LMs. In this work, we perform analysis using MedBERT and leave the choice of embedding to be determined by the multimodal data distribution and performance evaluation.
 Prioritizing fairness without considering performance will cause more adverse effects in healthcare and therefore a multi-criteria approach in decision-making is necessary \cite{baltussen2006priority}.

%#########

%###########

\subsection{Intersectional Bias Analysis and Mitigation}
\label{sec:bias}
\textbf{Bias analysis:} We perform a comprehensive analysis concerning multiple sensitive attributes, focusing on \textit{race}, \textit{gender}, and their intersectional subgroup pairs with algorithmic fairness metric, $\mathcal{U}(A, \hat{Y}) \forall A \in SG$, as detailed in Section \ref{sec:background}.

 Table \ref{tab:bias_mitigation} highlights limitations of how examining fairness metrics for a single sensitive attribute can obscure inequalities when intersectional subgroups are involved. The base classifier, $\mathcal M^{base}$ shows no significant disparity when evaluating the DP, TPR, and WP solely for gender. Conducting a more nuanced intersectional analysis on race and gender reveals that the disparity (DP) is even more pronounced in Asian Females than in Asian Males in ED Disposition.

 %This is not surprising given the skewed dataset distribution on the Asian categories. 
 
 %In effect, the WP of intersectional subgroups in both disposition and multitask classifications are even higher compared to their binary counterparts.
 \begin{figure*}[hbt!]
    \centering  \includegraphics[width=\textwidth,height=5cm]{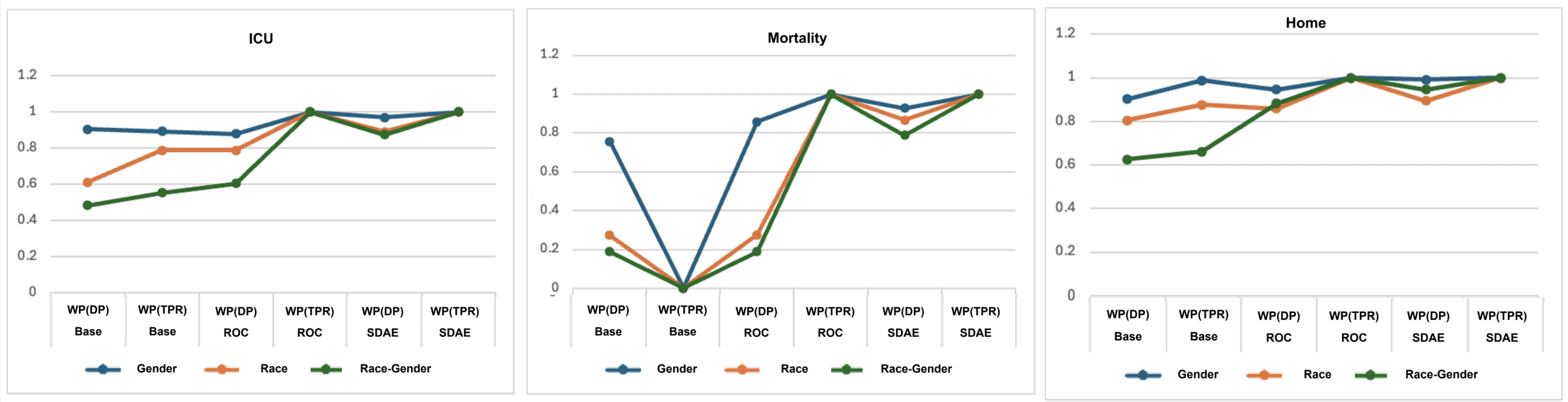} 
    \caption{Comparison of WP in multitask setting using MIMIC-Eye}
    \label{fig:wp}
\end{figure*}
\textbf{Bias mitigation:} Here, we compare our SDAE approach against state-of-the-art Reject Option Classification (ROC) \cite{kamiran2012decision}. Note that the ensemble, MAAT \cite{chen2022maat} relies on output probabilities which limits the applicability of post-processing methods. 
 We compare SDAE against (i) the Base classifier with ROC (gender), (ii) the Base classifier with ROC (race), and (iii) the base classifier with ROC (gender+race). The results are shown in Tables \ref{tab:bias_mitigation} and \ref{tab:bias_mitigation1} respectively for MIMIC-Eye and the biased sampling from MIMIC-IV ED. The main findings are: 
 \begin{enumerate}
     \item \textbf{Existing ROC approach injects additional biases at the intersections: } When performed ROC based only on gender, it injects additional biases into the race subgroups as the method tends to assign favorable outcomes to the gender and unfavorable outcomes to the subgroups of race and gender as both consist of privileged and unprivileged groups. This shows that when the intersectional subgroups are correlated, ROC on one sensitive attribute will have negative impacts on the intersectional groups (this holds for ROC on race as well).
     \item \textbf{Existing methods suffer from the problems of leveling down while SDAE improves the performance of minority:} When enforcing DP using ROC in MIMIC-eye, the value for the racial minority, \textit{Asian}, drops from 0.575 to 0.468. In contrast, our SDAE method improves it to 0.601, resulting in a positive movement for the minority group. In all cases from both MIMIC-Eye and MIMIC-IV ED datasets, our SDAE method enhances the performance of minority groups. ROC methods achieve fairness by reducing both majority and minority values, but this approach leads to degradation in minority performance which is precisely for whom the mitigation strategies are designed.
     \item \textbf{The existing ROC with race+gender has a null effect on intersectionality:} This proves the inability of single attribute settings to intersectionality.
     \item \textbf{The metric worst-case parity (WP) does not reflect the overall fairness reduction:} For the gender group in the MIMIC-Eye disposition task, there is a drop in DP values (.702 to .610) of ROC- models despite achieving a WP of 1. This shows that WP accounts only for the ratio, not the group-wise drop. 
\end{enumerate}
 \textit{Note:} The TPR for Female-Asian has been improved with SDAE, but not the DP as the negative probabilities lie close to $0$. We recommend choosing the threshold $\tau_{\text{SG}}$ (Equation 11) to force them to be in a positive class conditioned on an acute index of patients, as a lower-priority acute index does not demand an ED admission. If one wants them to be in a positive class, the threshold $\tau_{\text{SG}}$ can be set to a very low value for the subgroups, which we don't recommend as DP only considers positive predictions, not the ground truth. The overall WP for multitask comparison in MIMIC-Eye is given in Figure \ref{fig:wp} with results for MIMIC -IV ED in the supplementary file, Section 5. 
 
\begin{figure}[hbt!]
    \centering  \includegraphics[width=0.47\textwidth,height=3cm]{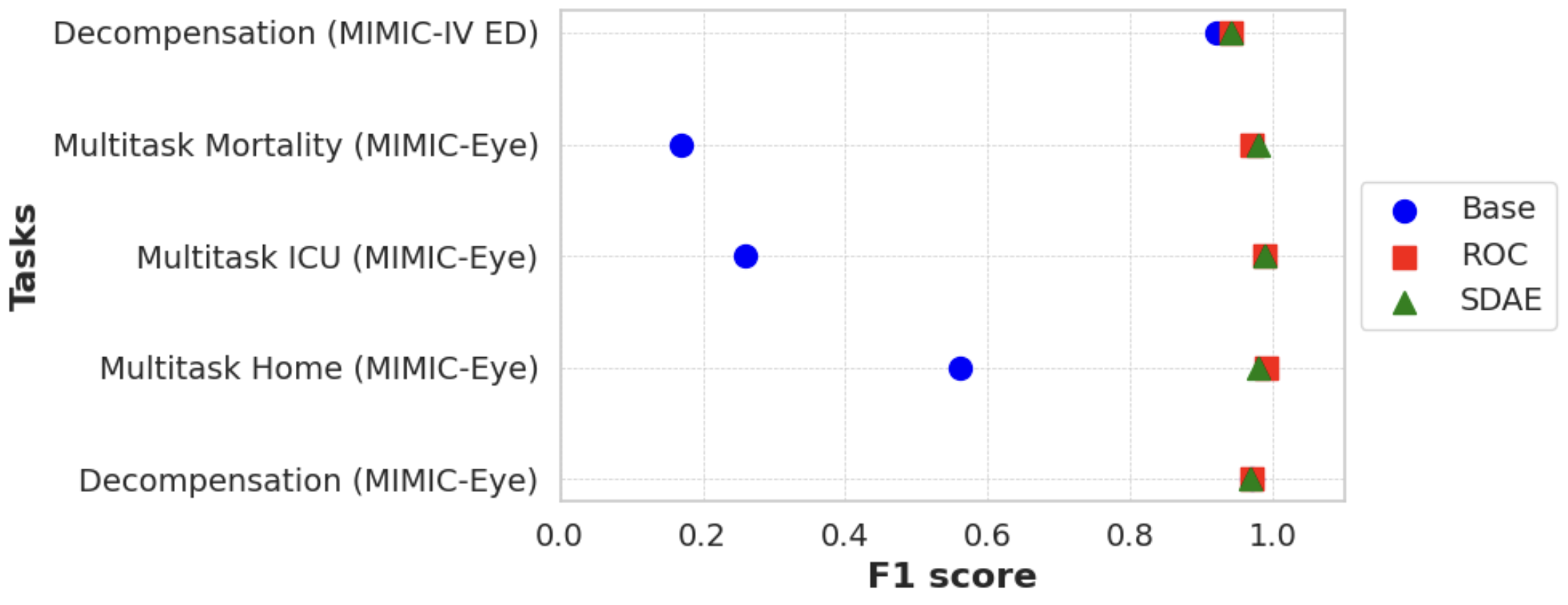} 
    \caption{Comparison of F1 scores}
    \label{fig:f1}
\end{figure}

\textbf{Significance of SDAE:} From Tables \ref{tab:bias_mitigation} and \ref{tab:bias_mitigation1}, and Figures \ref{fig:wp} and \ref{fig:f1} it is evident that our proposed SDAE model demonstrates consistent improvement in fairness (DP, TPR, and WP) across all tasks and datasets, with negligible performance degradation (Figure \ref{fig:f1}). The reason can be attributed to the unified text representation learning from modalities coupled with the joint advantages of ensembling and post-processing. The sub-group specific multimodal models are framed to capture the interoperability of features and exploit the disagreement. Unlike traditional ensembling, SDAE customizes the disagreement region to prevent identical scores across multiple subgroups, which effectively improves the fairness values of minorities.

%This contrasts with the ROC models and the base classifier. These results indicate that SDAE effectively leverages the disagreement region of subgroup ensembles to attribute predictions for fairness (DP or TPR) without causing minority performance degradation. 
 %The WP of Asian-females has not improved as all the models are in consensus to vote for negative class due to the low acute index value.  

\begin{table}[hbt!]
    \centering
    \small
    \setlength{\tabcolsep}{4pt}
    \renewcommand{\arraystretch}{1.2}
    \begin{tabular}{c|cc|cc|cc}
        \toprule
          \textbf{Group} & \multicolumn{2}{c|}{\textbf{$\mathcal{M}^{base}$}} & \multicolumn{2}{c|}{\textbf{ $\mathcal{M}^{ROC}$}} & \multicolumn{2}{c|}{\textbf{$\mathcal{M}^{SDAE}$}} \\
       \hline
         & DP & TPR & DP & TPR & DP & TPR \\
         \midrule
         Male(M) & \cellcolor{yellow!50}0.702 & \cellcolor{green!30}0.997 & \cellcolor{orange!30}{0.610$^-$}& \cellcolor{green!30}1.000 & \cellcolor{yellow!50}0.710 & \cellcolor{green!30}0.998 \\
         Female(F) & \cellcolor{yellow!50}0.703 & \cellcolor{green!30}0.998 & \cellcolor{orange!30}0.610$^-$ & \cellcolor{green!30}1.000 & \cellcolor{yellow!50}0.704 & \cellcolor{green!30}0.996 \\
        WP & \cellcolor{green!30}0.998 & \cellcolor{green!30}0.998 & \cellcolor{green!30}\textbf{1.000} & \cellcolor{green!30}\textbf{1.000} & \cellcolor{green!30}0.991 & \cellcolor{green!30}0.997 \\
        \hline
        White & \cellcolor{yellow!50}0.708 & \cellcolor{green!30}0.997 & \cellcolor{orange!30}0.611$^-$ & \cellcolor{green!30}1.000 & \cellcolor{yellow!50}0.702 & \cellcolor{green!30}0.998 \\
        Black & \cellcolor{yellow!50}0.707 & \cellcolor{green!30}0.996 & \cellcolor{orange!30}0.622$^-$ & \cellcolor{green!30}1.00 & \cellcolor{orange!30}0.688 & \cellcolor{green!30}0.998 \\
        Asian & \cellcolor{orange!50}0.575 & \cellcolor{green!30}1.000 & \cellcolor{orange!90}0.468$^-$ & \cellcolor{green!30}1.000 & \cellcolor{orange!30}0.601 & \cellcolor{green!30}0.998 \\
        WP & \cellcolor{green!30}0.812 & \cellcolor{green!30}0.996 & \cellcolor{yellow!50}0.752$^-$ & \cellcolor{green!30}\textbf{1.000} & \cellcolor{green!30}\textbf{0.856} & \cellcolor{green!30}\textbf{1.000} \\
        \hline
        White-M & \cellcolor{orange!30}0.697 & \cellcolor{green!30}0.995 & \cellcolor{orange!30}0.602$^-$ & \cellcolor{green!30}1.000 & \cellcolor{orange!30}0.687 & \cellcolor{green!30}1.000 \\
        White-F & \cellcolor{yellow!50}0.719 & \cellcolor{green!30}0.998 & \cellcolor{orange!30}0.620$^-$ & \cellcolor{green!30}1.000 & \cellcolor{yellow!50}0.709 & \cellcolor{green!30}1.000 \\
        Black-M & \cellcolor{yellow!50}0.720 & \cellcolor{green!30}0.996 & \cellcolor{orange!30}0.631$^-$ & \cellcolor{green!30}1.000 & \cellcolor{yellow!50}0.717 & \cellcolor{green!30}0.996 \\
        Black-F & \cellcolor{orange!30}0.696 & \cellcolor{green!30}0.997 & \cellcolor{orange!30}0.615$^-$ & \cellcolor{green!30}1.000 & \cellcolor{orange!30}0.688 & \cellcolor{green!30}0.996 \\
        Asian-M & \cellcolor{orange!30}0.638 & \cellcolor{green!30}1.000 & \cellcolor{orange!50}0.511$^-$ & \cellcolor{green!30}1.000 & \cellcolor{orange!30}0.638 & \cellcolor{green!30}1.000 \\
        Asian-F & \cellcolor{orange!50}0.511 & \cellcolor{green!30}1.000 & \cellcolor{orange!90}0.426$^-$ & \cellcolor{green!30}1.000 & \cellcolor{orange!50}0.510 & \cellcolor{green!30}1.000 \\
        WP & \cellcolor{yellow!50}0.709 & \cellcolor{green!30}0.995 & \cellcolor{orange!30}0.675$^-$ & \cellcolor{green!30}\textbf{1.000} & \cellcolor{yellow!50}\textbf{0.711} & \cellcolor{green!30}0.996 \\
        \bottomrule
    \end{tabular}
    \caption{Fairness comparison on MIMIC-Eye using the base classifier ($\mathcal M^{base}$), ROC models ($\mathcal M^{ROC}$), and the SDAE ( $\mathcal M^{SDAE}$) in Disposition. We enforced ROC on gender, race, and gender+race, and obtained identical results for all three ROC models, which we denote as $\mathcal{M}^{ROC}$ (best results are bolded) and a $'-'$ indicates the values are reduced from the base model (more than 5\%  reduction) though WP is improved. Here, DP is calculated based on the positive prediction of each subgroups and TPR is the true positive rate.}
    \label{tab:bias_mitigation}
\end{table}

\begin{table}[]
    \centering
    \small
    \setlength{\tabcolsep}{4pt}
    \renewcommand{\arraystretch}{1.2}
    \begin{tabular}{c|cc|cc|cc}
        \toprule
          \textbf{Group} & \multicolumn{2}{c|}{\textbf{$\mathcal{M}^{base}$}} & \multicolumn{2}{c|}{\textbf{$\mathcal{M}^{ROC}$}} & \multicolumn{2}{c|}{\textbf{$\mathcal{M}^{SDAE}$}} \\
       \hline
         & DP & TPR & DP & TPR & DP & TPR \\
         \midrule
         Male(M) & \cellcolor{orange!50}0.399 & \cellcolor{green!30}0.960 & \cellcolor{orange!70}0.370$^-$ & \cellcolor{green!30}1.000 & \cellcolor{orange!50}0.398 & \cellcolor{green!30}0.991 \\
    Female(F) & \cellcolor{orange!50}0.397 & \cellcolor{green!30}0.960 & \cellcolor{orange!70}0.366$^-$ & \cellcolor{green!30}1.000 & \cellcolor{orange!50}0.396 & \cellcolor{green!30}0.992 \\
    WP & \cellcolor{green!30}0.994 & \cellcolor{green!30}1.000 & \cellcolor{green!30}0.989 & \cellcolor{green!30}1.000 & \textbf{\cellcolor{green!30}0.994} & \cellcolor{green!30}0.998 \\
    \hline
    White & \cellcolor{orange!70}0.420 & \cellcolor{green!30}0.969 & \cellcolor{orange!70}0.397$^-$ & \cellcolor{green!30}1.000 & \cellcolor{orange!60}0.409 & \cellcolor{green!30}0.990 \\
    Non-White & \cellcolor{orange!80}0.339 & \cellcolor{green!30}0.928 & \cellcolor{orange!80}0.289$^-$ & \cellcolor{green!30}1.000 & \cellcolor{orange!70}0.363 & \cellcolor{green!30}1.000 \\
    WP & \cellcolor{green!30}0.807 & \cellcolor{green!30}0.957 & \cellcolor{yellow!60}0.727 & \textbf{\cellcolor{green!30}1.000} & \textbf{\cellcolor{green!30}0.887} & \textbf{\cellcolor{green!30}0.990} \\
    \hline
    White-M & \cellcolor{orange!70}0.420 & \cellcolor{green!30}0.969 & \cellcolor{orange!70}0.398$^-$ & \cellcolor{green!30}1.000 & \cellcolor{orange!60}0.409 & \cellcolor{green!30}0.990 \\
    White-F & \cellcolor{orange!70}0.421 & \cellcolor{green!30}0.969 & \cellcolor{orange!70}0.397$^-$ & \cellcolor{green!30}1.000 & \cellcolor{orange!60}0.410 & \cellcolor{green!30}0.989 \\
    NonWhite-M & \cellcolor{orange!80}0.345 & \cellcolor{green!30}0.925 & \cellcolor{orange!80}0.288$^-$ & \cellcolor{green!30}1.000 & \cellcolor{orange!70}0.367 & \cellcolor{green!30}1.000 \\
    NonWhite-F & \cellcolor{orange!80}0.341 & \cellcolor{green!30}0.932 & \cellcolor{orange!80}0.292$^-$ & \cellcolor{green!30}1.000 & \cellcolor{orange!70}0.361 & \cellcolor{green!30}1.000 \\
    WP & \cellcolor{green!30}0.809 & \cellcolor{green!30}0.954 & \cellcolor{yellow!60}0.723 & \textbf{\cellcolor{green!30}1.000} & \textbf{\cellcolor{green!30}0.880} & \cellcolor{green!30}0.989 \\
   
        \bottomrule
    \end{tabular}
    \caption{Fairness comparison between the base classifier, ROC  models, and the proposed SDAE in ED Disposition for a biased sampling in MIMIC-IV-ED dataset. The $'-'$ indicates group wise value reduction (more than 5 percent) and the proposed SDAE achieves DP and TPR without groupwise reduction. }
    \label{tab:bias_mitigation1}
\end{table}

\section{Related Works} 
There have been numerous pieces of research on biases in various clinical data sources and tasks \cite{pivovarov2014identifying, boyd2023equity, banerjee2023shortcuts, tripathi2023understanding, arias2023biases}. These studies have overlooked the biases in multimodal data sources with varying intensities and patterns. Recent works have shown efforts in identifying intersectional biases in clinical decision tasks \cite{ogungbe2019systematic, okoro2022intersectional}. However, to the best of our knowledge, there is no research in identifying and mitigating intersectional biases in tasks utilizing extensive multimodal data from all the sources, especially in the medical domain. Current intersectional bias mitigation strategies \cite{foulds2020bayesian, chen2022maat} cannot be extended to multimodality as these are tailored for their specific in-process/pre-process algorithms, limiting their flexibility to multimodal models.
\setlength{\parskip}{0pt}
\section{Conclusions and Future Works}
\setlength{\parskip}{0pt}
Our work introduced a novel framework, SDAE, designed for bias mitigation involving intersectional subgroups, utilizing multimodal information from clinical databases. We have shown that learning a unified feature representation in the form of text from multimodal data sources improves the overall performance of the model. Additionally, the modality-wise performance metrics illustrate that multimodal models positively contribute to the overall prediction performance in binary and multitask classifications. The bias analysis focused on the intersectional subgroup level revealed the inability of traditional single-attribute-centric approaches to capture the biases at those intersections. To account for this, we proposed a Subgroup-specific Discrimination-Aware Ensemble (SDAE), a post-processing method customized at the intersection level. The extensive intersectional fairness analysis on binary and multitask learning illustrated that the proposed SDAE improves the Worst case Demographic Parity of minorities when evaluated on single attribute and intersectional settings, without reduction in group-wise DP. This is in contrast to conventional methods, which have shown improvement in single attribute scenarios while sacrificing overall fairness values. Also, the Worst case True Positive Rate is reasonably maintained across the subgroups and achieved reasonable performance compared to Reject Option Classification (ROC) methods. Apart from the intersectional bias analysis, modality utilization, and bias mitigation, we benchmark the models and tasks presented in this work highlighting the challenges associated with diverse, complex, and data-rich modalities. We anticipate that future bias mitigation applied to MIMIC-Eye and MIMIC-IV ED will leverage our work, to encompass more tasks and analysis across patient visits.

 The future directions for this study will be to extend our SDAE framework to encompass mitigation using the entire pipeline including pre-processing and in-processing methods and framing a metric for measuring the intersectional biases reflecting the overall fairness change across the sub-groups. 

\section*{Limitations}
Despite the numerous advantages outlined in this work in terms of intersectional fairness and multimodal learning without performance degradation, there are limitations to our work. Our Subgroup-specific Discrimination Aware Ensembling (SDAE) will be computationally expensive if the sensitive attributes have high cardinality, resulting in a high number of intersections (in our empirical analysis we have used subgroup pairs up to 9). Also, our SDAE is designed as a post-process mitigation, and its adaptability to in-process and pre-process strategies has not been explored here. It is worth mentioning that MIMIC-Eye being a combination of MIMIC-IV and MIMIC-CXR has only 15627 stay IDs. We have overcome the limitations of small-scale data learning using clinical Language Models and showed that the performance has been reasonably maintained. But, in situations where pre-trained models are not available for applications, the multimodality information extraction will be cumbersome. 
%MIMIC-Eye is the only available dataset to the research community, having all the modalities including Xrays, which are overlooked in the literature. We, therefore, recommend using our method to foster future learning with the datasets in the extensive setting of intersectionality and multimodality. 
\section*{Ethics Statement}
The performance and bias metrics used in this work are taken from the literature and not from any clinical trials or real-world scenarios. Therefore before using our work in real-world settings, it is advisable to get recommendations from healthcare experts, ethicists, and policy makers. Another important point is that the major sources of biases in this work depend on the distribution of demographic data, model selection, and modality selection. We, therefore, recommend conducting an extensive bias analysis on any real-world dataset before generalizing our method. Finally, we advocate for analyzing group-wise performance drops in fair interventions despite satisfying the overall fairness criteria as interventions can be explicitly fair. This causes performance drops in minority, for whom we aim for fairness enforcement. 
%The problems of leveling down should be analyzed and quantified to ensure equity and resolve disparity, especially in healthcare settings.

%\section*{Acknowledgements}


\bibliography{aaai25}

\section{Reproducibility Checklist}

This paper:
\begin{enumerate}
    \item Includes a conceptual outline and/or pseudocode description of AI methods introduced (yes)

\item Clearly delineates statements that are opinions, hypothesis, and speculation from objective facts and results (yes)
\item Provides well marked pedagogical references for less-familiare readers to gain background necessary to replicate the paper (yes)
\item Does this paper make theoretical contributions? (no)

\item A motivation is given for why the experiments are conducted on the selected datasets (yes)
\item All novel datasets introduced in this paper are included in a data appendix. (yes/links are given)
\item All novel datasets introduced in this paper will be made publicly available upon publication of the paper with a license that allows free usage for research purposes. (NA)
\item All datasets drawn from the existing literature (potentially including authors’ own previously published work) are accompanied by appropriate citations. (yes)
\item All datasets drawn from the existing literature (potentially including authors’ own previously published work) are publicly available. (partial/have to complete training)
\item All datasets that are not publicly available are described in detail, with explanation why publicly available alternatives are not scientifically satisficing. (yes)
\item Does this paper include computational experiments? (yes)

If yes, please complete the list below.

\item Any code required for pre-processing data is included in the appendix. (yes).
\item All source code required for conducting and analyzing the experiments is included in a code appendix. (yes)
\item All source code required for conducting and analyzing the experiments will be made publicly available upon publication of the paper with a license that allows free usage for research purposes. (yes)
\item All source code implementing new methods have comments detailing the implementation, with references to the paper where each step comes from (yes)
\item If an algorithm depends on randomness, then the method used for setting seeds is described in a way sufficient to allow replication of results. (yes)
\item This paper specifies the computing infrastructure used for running experiments (hardware and software), including GPU/CPU models; amount of memory; operating system; names and versions of relevant software libraries and frameworks. (yes)
\item This paper formally describes evaluation metrics used and explains the motivation for choosing these metrics. (yes)
\item This paper states the number of algorithm runs used to compute each reported result. (yes)
\item Analysis of experiments goes beyond single-dimensional summaries of performance (e.g., average; median) to include measures of variation, confidence, or other distributional information. (yes)
\item The significance of any improvement or decrease in performance is judged using appropriate statistical tests (e.g., Wilcoxon signed-rank). (yes)
\item This paper lists all final (hyper-)parameters used for each model/algorithm in the paper’s experiments. (yes)
\item This paper states the number and range of values tried per (hyper-) parameter during development of the paper, along with the criterion used for selecting the final parameter setting. (yes)

\end{enumerate}

\end{document}